\documentclass{article}
\usepackage{harvard}
\usepackage{proof}
\usepackage{amssymb}
\usepackage{latexsym}
\usepackage{amsmath}
\usepackage{color}
\usepackage{linguex}

\title{Chart Parsing Multimodal Grammars}
\author{Richard Moot}
\date{University of Montpellier, CNRS (LIRMM)}

\definecolor{fade}{gray}{.6}
\newcommand{\emptylist}{[]}

\begin{document}
\maketitle

\section{Introduction}

The short note describes the chart parser for multimodal categorial
grammars which has been developed in conjunction with the type-logical
treebank for French, which is described in more detail in
\cite{moot10semi,moot12spatio,moot15tlgbank} and which is available at \cite{tlgbank}. 
The chart parser itself can be downloaded as a part of Grail light at \cite{graillight}.

\medskip
\texttt{https://github.com/RichardMoot/GrailLight}
\medskip

The chart parser is an instance of the ``deductive parsing''
technology of \cite{dedpar} and the core parsing engine of their
implementation has been retained in the source coude, with only some
minor modifications. I am grateful to the authors for
having made their source code available.


The current chart parser was originally introduced as a preprocessing
step for a proof net algorithm \cite{moot17grail}. However, this preprocessing step turned out to
be so effective that it soon handled a bit under 98\% of the complete
French Type-Logical Treebank and therefore it made sense to add
additional chart rules to handle the remaining few percent as well
(these are briefly sketched in Section~\ref{sec:other}, the rest of
this paper focuses on the basic rules of the chart parser).


This paper presupposes the reader has at least a basic familiarity
with multimodal categorial grammars \cite{m10sep,Moo11,mr12lcg} and
with chart parsing \cite{dedpar}. 

\section{Chart rules}

In this section, I will dicuss the inference rules used by the chart
parser. I will start with the simplest rules and gradually introduce
more detail.

\subsection{AB rules}
\label{sec:ab}

The elimination rules $/E$ and $\backslash E$ appear already in \cite{dedpar}.
For AB grammars, the chart rules are very simple and shown in
Figure~\ref{fig:abchart}. Chart items are tuples $\langle \Gamma, F, L,
R\rangle$ where $\Gamma$ is an antecedent term, $F$ is a formula, and $L$
and $R$ are integers representing the
leftmost and rightmost string positions respectively. The meaning of such a tuple is that we
have derived a formula $F$, using the hypotheses in $\Gamma$, spanning exactly the positions from
$L$ on the left to $R$ on the right.\footnote{The use of pairs of string positions to represent
  substrings of an input string is widely used in
  parsing algorithms; see for example \cite{PS87,dedpar,jm}.}

\begin{figure}
$$
\begin{array}{ccc}
\infer[/E]{\langle \Gamma\circ \Delta, A, I, K \rangle}{\langle
  \Gamma, A/B, I, J \rangle & \langle \Delta, B,J,K \rangle}
&&
\infer[\backslash E]{\langle \Gamma\circ \Delta, A, I, K
  \rangle}{\langle \Gamma, B, I, J \rangle & \langle \Delta, B\backslash A,J,K \rangle}
\\
\end{array}
$$
\caption{AB grammar chart rules}
\label{fig:abchart}
\end{figure}

With this in mind, the chart rule for $/E$ indicates that if we have
derived a a formula $A/B$ spanning string positions $I-J$ and a
formula $B$ spanning string positions $J-K$ (that is, $A/B$ and $B$
are adjacent with $B$ immediately to the right of $A/B$), then we can
conclude that we can derive a constituent $A$ from positions $I$ to
$K$ (that is, the concatenation of the strings assigned to $A/B$ and
$B$).

Given these rules, proving an AB sequent $A_1,\ldots,A_n\vdash B$
corresponds to starting from axioms $\langle w_1, A_1, 0, 1\rangle \ldots
\langle w_n, A_n, n-1, n\rangle$ and deriving the goal $\langle \Gamma, B,
0, N\rangle$ with $\textit{yield}(\Gamma)= w_1,\ldots,w_n$. To
facilitate inspection of the chart items, $\Gamma$ will not be a
binary tree of \emph{formulas}, but a binary tree of the corresponding
\emph{words}. Therefore, a lexical entry for the verb ``dort''
(\textit{sleeps}) with formula $np\backslash s$ at
position 1-2 will correspond not to the item $\langle np\backslash
s, np\backslash s, 1, 2\rangle$ but to the item $\langle
\textit{dort}, np\backslash s, 1, 2\rangle$.

\paragraph{Example} As an example, the table below shows how the chart
is filled for ``Le march\'{e} financier de Paris'' (\emph{the financial
market of Paris}).

{
\newcommand{\wa}{\textit{le}}
\newcommand{\wb}{\textit{march\'{e}}}
\newcommand{\wc}{\textit{financier}}
\newcommand{\wrd}{\textit{de}}
\newcommand{\we}{\textit{Paris}}
\medskip

\begin{tabular}{r|ll}
 & Chart Item & Justification \\ \hline 
1 & $\langle \wa, np/n, 0, 1 \rangle$ & Lexicon \\
2 & $\langle \wb, n, 1, 2 \rangle$ & Lexicon \\
3 & $\langle \wc, n\backslash n, 2, 3 \rangle$ & Lexicon \\
4 & $\langle \wrd, (n\backslash n)/np, 3, 4 \rangle$ & Lexicon \\
5 & $\langle \we, np, 4, 5 \rangle$ & Lexicon \\
\textcolor{fade}{6} & \textcolor{fade}{$\langle \wa\circ\wb, np, 0, 2 \rangle$} & \textcolor{fade}{From
  1,2 by $/E$} \\
7 & $\langle \wb\circ\wc, n, 1, 3 \rangle$ & From 2,3 by $\backslash E$\\
8 & $\langle \wrd\circ\we, n\backslash n, 3, 5\rangle$ & From 4,5 by $/ E$ \\
\textcolor{fade}{9} & \textcolor{fade}{$\langle \wa\circ(\wb\circ\wc), np, 0, 3 \rangle$} & \textcolor{fade}{From 1,7 by $/ E$} \\
10 & $\langle (\wb\circ\wc)\circ(\wrd\circ\we), n, 1, 5 \rangle$ & From 7,8 by $\backslash E$\\
11 & $\langle \wa\circ((\wb\circ\wc)\circ(\wrd\circ\we)), np, 0, 5 \rangle$ & From 1,10 by $/ E$ \\
\end{tabular}
}
\medskip

The chart items are labeled from 1 to 11 indicating the order they are
entered in the chart. We use a general chart parser of the type
explained in \cite{dedpar}, so we start with an agenda
containing items 1-5 (the lexical lookup for the words in the
sentence) and then successively add the items of the agenda to the
chart. When we add an item from the agenda to the chart, we compute
all consequences according to the rules of the grammar of this item
with all items already in the chart. So once item 2 is added to the
chart, item 6 is added to the agenda, since it is the combination of
item 2 with item 1 (already in the chart) by means of rule
$/E$. Similarly, item 7 is added to the agenda when item 3 is added to
the chart and item 8 is added to the agenda when item 5 is added to
the chart, etc.

We complete the parse when item 11 is added to the chart. If desired,
we can recover the proof by recursively finding the justification of
each of the rules, going back from 11 to 1 and 10, from 10 to 7 and 8
(1 is in the lexicon and so an axiom of the proof) until we have
reached all the axioms, which are justified by their respective
lexical entries. The chart items marked in gray do not contribute to
the proof of 11.

\paragraph{Implementation notes} The actual implementation keeps track
of several types of additional information: it computes the semantics of the
derivation and there is also a mechanism for computing the
(log-)probabilities of the rules.

The implementation also uses an important simplification: once we have
computed a chart item for a formula $A$ over span $I$ and $J$ then we
will treat this as known and reject any further derivations of this
formula $A$ over the same string (if probabilities are used, only the
most probably derivation of $A$ is kept). This can throw away alternative
semantic readings for a phrase, but reduces the size of the chart.
If desired, this behavior can easily by changed by replacing the don't care variables
\verb+_+ in the predicate \verb+subsumes_data+ by a test for
$\alpha$-equivalence of the lambda-terms.

\subsection{Hypothetical Reasoning}

Hypothetical reasoning is implemented using a strategy very similar
to ``gap threading'' in the parsing literature. Chart items are now of the form $\langle \Gamma, F, L,
R, e\rangle$, where $e$ is a set of pairs of the form $P-A$, with $P$
a position integer and $A$ a formula; the set $e$ is the set of
``extracted'' constituents which have been used to compute $F$. The rules for extraction (hypothetical
reasoning) are shown in Figure~\ref{fig:hr}. The $/E$ and $\backslash
E$ rules of Figure~\ref{fig:abchart} have been updated to include the
new set $e$ of extracted items.

The set union $e_1\cup e_2$ of two such sets $e_1$ and $e_2$ is defined only if $e_1\cap
e_2$ is empty; this reflects that fact that a hypothesis to be
discharged later can only be used once.

The $\textit{e\_start}$ rule states that if we have a formula
$X/(Y/\Diamond_1\Box_1 B)$ with rightmost position $K$ and a formula
$A/B$ spanning positions $I-J$ then we can conclude there is a formula
$A$ spanning positions $I-J$ depending on an extracted element
$K-B$. The underscores $\_$ indicate we do note care about this value
for the chart item. So for the leftmost premiss of the
$\textit{e\_start}$, we do not care about the antecedent, about
the leftmost position or about the stack of extractions: the formula
$X/(Y/\Diamond_1\Box_1 B)$ functions as a sort of ``trigger'' allowing
extraction of a $B$ formula to take place to its right.

The rule $\textit{e\_start}$ has as side conditions that $K \leq I$ and
that $K-B$ is not a member of $e$ (this is a general consequence of the
disjoint set union used).

The $\textit{e\_start}$ rule is a combination of using a $B\vdash B$
axiom in combination with a previous proof of $\Gamma \vdash A/B$ 
to derive by $/ E$ that $\Gamma,B \vdash A$, with the condition that
the $B\vdash B$ hypothesis must be discharged at position $K$ by the
formula $X/(Y/\Diamond_1\Box_1 B)$ which licensed this rule. This
discharge is taken care of by the $\textit{e\_end}$  rule.

The $\textit{e\_end}$ rule states that if we have derived a $Y$ using
a hypothetical $B$ to the immediate right of a formula
$X/(Y/\Diamond_1\Box_1 B)$, then we can derive an $X$ spanning the
total positions, removing the formula $J-B$ from the set of extracted elements; the
notation $e_2 \cup \{ J-B\}$ indicates that the $Y$ formula was
derived using the formula $B$ exactly once (plus some additional,
possibly empty, set of items $e_2$).

 A chart item is \emph{coherent},
if for all $P-A \in e$, $L \leq P$. This is because formulas of the
form $X/(Y/\Diamond_1\Box_1 B)$ are looking to their right for a
constituent $Y$ missing a $B$ somewhere.

We initialize all lexical entries with the empty set and at the end of a derivation, we require that the set of traces is
empty. That is, our lexical entries are now of the form $\langle
w_i,A_i,i-1,i,\emptyset\rangle$ and our goal is of the form $\langle \Gamma, C,
0, N, \emptyset \rangle$ for some formula $C$ and with the antecedent
term $\Gamma$ such that $\textit{yield}(\Gamma)= w_1,\ldots,w_n$.

\begin{figure}
$$
\begin{array}{c}
\infer[/E]{\langle \Gamma\circ \Delta, A, I, K, e_1\cup e_2
  \rangle}{\langle \Gamma, A/B, I, J, e_1
  \rangle & \langle \Delta, B,J,K, e_2 \rangle}
\\
\\
\infer[\backslash E]{\langle \Gamma\circ\Delta, A, I, K, e_1 \cup e_2
  \rangle}{\langle \Gamma, B,
  I, J, e_1 \rangle & \langle \Delta, B\backslash A,J,K, e_2 \rangle}
\\
\\
\infer[\textit{e\_start}]{\langle \Gamma, A, I, J, e\cup
  \{K-B\}\rangle}{\langle \_, X/(Y/\Diamond_1\Box_1 B), \_, K, \_
  \rangle & \langle \Gamma, A/B, I, J, e \rangle }
\\ \\
\infer[\textit{e\_end}]{\langle \Gamma\circ\Delta, X, I, K, e_1 \cup e_2 \rangle}{\langle
  \Gamma, X/(Y/\Diamond_1\Box_1 B), I, J, e_1 \rangle & \langle
  \Delta, Y,J,K, e_2 \cup \{ J-B \} \rangle}
\\
\end{array}
$$
\caption{Hypothetical reasoning chart rules}
\label{fig:hr}
\end{figure}


Typical instantiations of the formula $X/(Y/\Diamond_1\Box_1 B)$ are
$(n\backslash n)/(s/\Diamond_1\Box_1 np)$ (for relativizers) and
$(np\backslash s)/((np\backslash s)/\Diamond_1\Box_1 np)$ (for clitics).

\paragraph{Example} The chart rules for extraction/hypothetical
reasoning are perhaps the easiest to understand by seeing them in
action. We can derive the sentence fragment ``qu'on emprunte''
(\emph{that we borrow}) to be of type $n\backslash n$ as follows.

{
\newcommand{\wa}{\textit{qu'}}
\newcommand{\wb}{\textit{on}}
\newcommand{\wc}{\textit{emprunte}}
\begin{tabular}{r|ll}
 & Chart Item & Justification \\ \hline
1 &$\langle \wa,(n\backslash n)/(s/\Diamond_1\Box_1 np), 2,3, \emptyset
\rangle$ & Lex \\
2 & $\langle \wb,np, 3, 4, \emptyset \rangle$ & Lex \\
3 & $\langle \wc,(np\backslash s)/np, 4, 5, \emptyset \rangle$ & Lex \\
4 & $\langle \wc,np\backslash s,4,5, \{ 3-np \} \rangle$ & 1,3
$\textit{e\_start}$ \\
5 & $\langle \wb\circ\wc,s, 3, 5, \emptyset \cup \{3-np\} \rangle$ & 2,4
$\backslash E$ \\
6 & $\langle \wa\circ(\wb\circ\wc), n\backslash n, 2, 5, \emptyset \rangle$ & 1,5
$\textit{e\_end}$ \\
\end{tabular}
}

\paragraph{Incompleteness of the rules} As can be seen from the rules, they are
incomplete. The \emph{extraction start} rule can apply only to
formulas of the form $X/(Y/\Diamond_1\Box_1 B)$, with a fixed
combination of implications (excluding, for example
$(Y/\Diamond_1\Box_1 B)\backslash X$ or $X/(Y\bullet\Diamond_1\Box_1
B)$ and only when the $B$ formula is an argument, since the
\emph{extraction start} rule is essentially the $/E$ rule applied to a
$B$ hypothesis ``at a distance''. Another restriction is that each
combination of rightmost position and extracted formula $R-B$ can
introduce only one hypothetical item. We would need additional chart rules
if we want to treat these other cases. The treatment of gapping,
discussed briefly in Section~\ref{sec:other}, allows the extracted
element to be the functor of an elimination rule.

Though this formula restriction and the resulting incompleteness are
unfortunate, since it requires us to be careful in case the algorithm doesn't
find a proof, this rule captures most of the occurrences of the
$\Diamond_1\Box_1$ mixed associativity/commutativity rather nicely. 

\paragraph{Implementation notes} The actual implementation also keeps
track of the  rightmost position $J$ used for the $\textit{e\_start}$
rule. So the set of items $e$ takes the form triples $K-J-B$ where $K$
is the rightmost position of the licensor formula and $J$ is the
rightmost position of the extracted $B$ formula. This allows us to use
a single rule schema for a combination of mixed associativity and
mixed commutativity --- the rules for $\Diamond_1\Box_1$ shown --- and
for $\Diamond_0\Box_0$ which only allow mixed associativity (or
``right-node raising''). The $\textit{e\_end}$ rule in this case
requires that the rightmost position $K$ of the constituent $Y$ is
also the rightmost position of the extracted $B$ formula. This
right-node raising analysis also has a rule for formulas of the form
$(Y/\Diamond_0\Box_0 B)\backslash X$ and can therefore treat lexical
formulas such as $((((np\backslash s)/\Diamond_0\Box_0 np)\backslash
(np\backslash s)/np))/((np\backslash s)/\Diamond_0\Box_0 np)$, which is a transitive verb
conjunction type but which allows combinations such as the following.

$$(np\backslash s)/(np\backslash s) , (np\backslash s)/np \vdash
(np\backslash s)/\Diamond_0\Box_0 np$$

This is useful for patterns like ``has read and might implement
(Dijkstra's algorithm'' , where both ``has read'' and ``might
implement'' require the derivation pattern shown above.

\subsection{Head wrap}

French adverbs can occur at the start of the sentence, at the end of
the sentence and before the verb (where we can assign them the formulas
$s/s$, $s\backslash s$ and $(np\backslash s)/(np\backslash s)$
respectively.\footnote{We have chosen an event semantics in the style
  of Davidson for adverbs, which means that we can treat many adverbs
  as sentence modifiers. Some subject-oriented adverbs, such as
  ``ensemble'' (\emph{together}) need both the subject $np$ and the
  sentence for their semantics and are assigned $(s/(np\backslash
  s))/np$ and $(np\backslash s)\backslash (np\backslash s)$ instead.}
In addition, French adverbs can occur directly after the verb but also
between a verb and its arguments. In order to avoid unnecessary
duplication in the lexicon, we assign adverbs the type $s\backslash_1 s$ (or,
in some cases, $(np\backslash s)\backslash_1 (np\backslash s)$) and
use structural rules to move the verb to a sentence-final position.

In Figure~\ref{fig:wrap} we see how this idea translates into chart
rules. In addition to the set of extracted items, our chart items now
contain a \emph{stack} of head-wrapped elements. We have chosen a
stack instead of a set here to avoid generating readings which would
correspond to permutations of the adverbs. With few exceptions,
adverbs take scope from left to right. In the chart rules, ``$+$''
corresponds to stack concatenation, $[H|T]$ indicates a stack with
first element $H$ and rest of the stack $T$ (which is itself a valid
stack) and $\emptylist$ is the empty stack.
We both end and start our proof with empty stacks ($h=\emptylist$) and empty sets of
traces ($e=\emptyset$). That is, our lexical entries are of the form $\langle w_i,
A_i, i-1, i, \emptyset,\emptylist\rangle$ and the  goal is $\langle \Gamma, B,
0, N, \emptyset, \emptylist \rangle$ with $\textit{yield}(\Gamma)= w_1,\ldots,w_n$

The \textit{wr} rule wraps a chart entry with formula $X\backslash_1
X$ to its correct \emph{syntactic} position, but also pushes it onto
the stack $h_2$. As can been seen from the rule, the stack $h_1$ is
then prefixed to this new stack, thereby keeping all the stack
elements in the desired order: the elements in $h_1$ before the new
item and the elements in $h_2$ after it.

Finally, the \textit{wpop} rule simply allows us to pop a stack
element $X\backslash_1 X$ whenever the current chart item containing
the stack is of type $X$.

\begin{figure}
$$
\begin{array}{c}
\infer[/E]{\langle \Gamma\circ \Delta, A, I, K, e_1\cup e_2, h_1+h_2 \rangle}{\langle
  \Gamma, A/B, I, J, e_1, h_1
  \rangle & \langle \Delta, B,J,K, e_2, h_2 \rangle}
\\
\\
\infer[\backslash E]{\langle \Gamma\circ\Delta, A, I, K, e_1 \cup e_2,
  h_1 + h_2 \rangle}{\langle \Gamma, B,
  I, J, e_1, h_1 \rangle & \langle \Delta, B\backslash A,J,K, e_2, h_2 \rangle}
\\
\\
\infer[\textit{e\_start}]{\langle \Gamma, A, I, J, e\cup
  \{K-B\}, h\rangle}{\langle \_, X/(Y/\Diamond_1\Box_1 B), \_, K, \_, \_ \rangle
  & \langle \Gamma, A/B, I, J, e, h \rangle }
\\ \\
\infer[\textit{e\_end}]{\langle \Gamma\circ \Delta, X, I, K, e_1 \cup e_2, h
  \rangle}{\langle \Gamma, X/(Y/\Diamond_1\Box_1 B), I, J, e_1, h \rangle &
  \langle \Delta, Y,J,K, e_2 \cup \{ J-B \}, \emptylist \rangle}
\\
\\
\infer[\textit{wr}]{\langle \Gamma\circ_1 \Delta, X, I, K, e_1\cup
  e_2, h_1+[\text{J-K}-Y\backslash_1 Y|h_2]\rangle}{\langle \Gamma, X, I, J, e_1, h_1\rangle & \langle
  \Delta, Y\backslash_1 Y, J, K, e_2, h_2\rangle}
\\
\\
\infer[\textit{wpop}]{\langle \Gamma, X, I, J, e, h\rangle}{\langle
  \Gamma, X, I, J, e, [\text{K-L}-X\backslash_1 X|h]\rangle}
\end{array}
$$
\caption{Head wrap chart rules}
\label{fig:wrap}
\end{figure}

\paragraph{Example} The wrapping rules are best illustrated by
example. The sentence ``il occupera ensuite diverses fonctions''
(\emph{he will occupy various functions afterwards}) is analysed as follows.

{
\newcommand{\wa}{\textit{il}}
\newcommand{\wb}{\textit{occupera}}
\newcommand{\wc}{\textit{ensuite}}
\newcommand{\we}{\textit{diverses}}
\newcommand{\wf}{\textit{fonctions}}
\begin{tabular}{r|ll}
 & Chart Item & Just.{} \\ \hline
1 &$\langle \wa, np, 0,1,
\emptyset, \emptylist
\rangle$ & Lex \\
2 & $\langle \wb,(np\backslash s)/np, 1, 2, \emptyset, \emptylist \rangle$ & Lex \\
3 & $\langle \wc, s\backslash_1 s, 2, 3, \emptyset, \emptylist\rangle$ & Lex\\
4 & $\langle \we,np/n, 3, 4, \emptyset, \emptylist\rangle$ &  
Lex \\  
5 & $\langle \wf,n, 4, 5, \emptyset, \emptylist\rangle$ &  
Lex \\  
6 & $\langle \wb\circ_1\wc,(np\backslash s)/np, 1, 3, \emptyset,
[\text{2,3}-s\backslash_1 s]\rangle$ & 2,3 $\textit{wr}$ \\ 
7 & $\langle \we\circ\wf, np, 3, 5, \emptyset, \emptylist\rangle$ &
4,5 $/ E$ \\
8 & $\langle (\wb\circ_1\wc)\circ(\we\circ\wf), np\backslash s, 1, 5, \emptyset,
[\text{2,3}-s\backslash_1 s]\rangle$ & 6,7 $/ E$ \\
9 & $\langle \wa\circ((\wb\circ_1\wc)\circ(\we\circ\wf)), s, 0, 5, \emptyset,
[\text{2,3}-s\backslash_1 s]\rangle$ & 1,8 $\backslash E$ \\
10 & $\langle \wa\circ((\wb\circ_1\wc)\circ(\we\circ\wf)), s, 0, 5, \emptyset,
\emptylist\rangle$ & 9 $\textit{wpop}$ \\
\end{tabular}
}

The parse first combines the transitive verb ``occupera'' (\emph{will
  occupy}, chart item 2) with the
adverb ``ensuite'' (\emph{afterwards}, chart item 3) by pushing the
adverb on the stack and by combining the lexical strings, producing
chart item 6. We continue the proof with elimination rules until we
derive $s$ from positions 0 to 5 but with the adverb still on the
stack. Since $s$ and $s\backslash_1 s$ match the formulas of a \emph{wpop} rule, we pop the adverb from the stack and produce the final
item 10. 

The example below shows the interaction of the head wrap and the
extraction rules.

\medskip
{
\newcommand{\wz}{\textit{qu'}}
\newcommand{\wa}{\textit{il}}
\newcommand{\wb}{\textit{occupera}}
\newcommand{\wc}{\textit{ensuite}}
\begin{tabular}{r|ll}
 & Chart Item & Just.{} \\ \hline
1 &$\langle \wz, (n\backslash n)/(s/\Diamond_1\Box_1 np), 0, 1,
\emptyset, \emptylist\rangle$ & Lex \\
2 &$\langle \wa, np, 1, 2,
\emptyset, \emptylist
\rangle$ & Lex \\
3 & $\langle \wb,(np\backslash s)/np, 2, 3, \emptyset, \emptylist \rangle$ & Lex \\
4 & $\langle \wc, s\backslash_1 s, 3, 4, \emptyset, \emptylist\rangle$ & Lex\\
5 & $\langle \wb, np\backslash s, 2, 3, \{ 1-np \}, \emptylist
\rangle$ &
1,3 $\textit{e\_start}$ \\
6 & $\langle \wb\circ_1\wc,(np\backslash s)/np, 2, 4, \emptyset,
[\text{2,3}-s\backslash_1 s]\rangle$ & 3,4 $\textit{wr}$ \\  
7 & $\langle\wa\circ\wb, s, 1, 3, \{ 1-np \}, \emptylist \rangle$ & 2,5
$\backslash E$ \\ 
8 & $\langle \wb\circ_1\wc, np\backslash s, 2, 4, \{ 1-np \},
[\text{2,3}-s\backslash_1 s]\rangle$ & 4,5 $\textit{wr}$ \\  
9 & $\langle (\wa\circ\wb)\circ_1 \wc, s, 1, 4, \{ 1-np \}, [\text{2,3}-s\backslash_1 s] \rangle$ & 4,7
$\textit{wr}$ \\
10 & $\langle (\wa\circ\wb)\circ_1 \wc, s, 1, 4, \{ 1-np \}, \emptylist \rangle$ & 9
$\textit{wpop}$ \\
11 & $\langle \wz \circ ((\wa\circ\wb)\circ_1 \wc), n\backslash n, 0, 4, \emptyset, \emptylist \rangle$ & 1,10
$\textit{e\_end}$ \\
\end{tabular}
}
\medskip

Using chart items 2 and 8 above, we could have applied the $\backslash E$ rule to
produce $\textit{il} \circ (\textit{occupera} \circ_1
\textit{ensuite})$, resulting in a chart item which would otherwise be
identical to item 9. Therefore, according the the implementation note
discussed at the end of Section~\ref{sec:ab}, this entry is treated as
``already known'' and not entered in the chart. Other chart items have
multiple equivalent derivations (including even the antecedent term):
for example, as shown in the table above, chart item 8 has been
derived from 4 and 5 using \textit{wr} but it has
an alternative derivation from 1 and 6 using \textit{e\_start}: there
are two equivalent ways to apply \textit{e\_start} and \textit{wr} to
the transitive verb to produce chart item 8.

Since the \textit{e\_end} rule requires an empty stack to apply, we
cannot apply the \textit{e\_end} rule to chart item 9 and need to pop
the stack first using \textit{wpop}, producing chart item 10, which is
the proper configuration for an application of \textit{e\_end}.

\paragraph{Implementation details} The implementation allows us to pop
$s\backslash_1s$ elements from the stack at the $np\backslash s$ level
as well. This allows infinitive arguments to take adverbs of the form
$s\backslash_1 s$.

\subsection{Other chart rules}
\label{sec:other}

\paragraph{Quoted speech} In newspaper articles, quotes speech is
rather frequent. Most frequently, this takes the form of a tag like
``said the Prime Minister'', and this does not necessarily occur
at the end of a sentence. To complicate matters, we even
have sentences like the following.

\exig. \label{s:qb}  [sl Les conservateurs], a ajout\'{e} le premier ministre ...,  [sr ``ne sont pas des
opportunistes qui virevoltent d'une politique \`{a} l'autre ] \\
{} [sl The {Conservatives}], has added the Prime Minister ..., [sr ``{} are not {}
  opportunists who flip-flop {from one} policy to another ]\\

In this sentence the quoted sentence is split into two parts (marked
$sl$ and $sr$) and there two parts together are the arguments of the
past participle ``ajout\'{e}'' (\emph{added}), which itself is the
argument of the auxiliary verb form ``a'' (\emph{has}) (and the elided
material ``...'' includes an adverb modifying the past participle).

As a solution, the additional chart rules treat these combinations
much like complex adverbs. For example, we can derive ``a ajout\'{e}
to be for type $s\backslash_1 s$ as follows.

$$
\infer[\backslash I]{(\textit{a} \circ \textit{ajout\'{e}}) \circ np \vdash
  s\backslash_1 s}{\infer[MA_1]{x \circ_1 ((\textit{a} \circ
    \textit{ajout\'{e}}) \circ np) \vdash s}{\infer[MC_1]{
    ( x \circ_1 (\textit{a} \circ
    \textit{ajout\'{e}})) \circ np\vdash s
}{\infer[/E]{(\textit{a}\circ (x \circ_1
    \textit{ajout\'{e}})) \circ np \vdash
    s}{\infer[/E]{\textit{a} \circ (x \circ_1 \textit{ajout\'{e}})
      \vdash s/np}{\infer[\textit{Lex}]{(s/np)/(np\backslash s_{\textit{ppart}})}{\textit{a}}
    & \infer[\backslash E]{x \circ_1 \textit{ajout\'{e}} \vdash np\backslash
      s_{\textit{ppart}}}{
      \infer[\textit{Hyp}]{x\vdash s}{} &
      \infer[\textit{Lex}]{s\backslash_1 (np\backslash s_{\textit{ppart}})}{\textit{ajout\'{e}}}
}} &
& \infer{np\vdash np}{}
}}}}
$$

\paragraph{Gapping} Gapping includes cases like those shown below. 

\exig. Le v\'{e}hicule pourrait \^{e}tre immobilis\'{e} et {la carte
grise} retenue. \\
 The car could be immobilised and {the registration certificate}
 retained \\

This sentence can be paraphrased along the lines ``the car could be
immobilised and the registration certificate \emph{could be}
retained'', with the verb group ``pourrait \^{e}tre'' (\emph{could
  be}) occurring only in the first sentence syntactically, but
semantically it fills the same role in both sentences. This type of
sentences is treated along the lines of \cite{cgellipsis}, though
recast in the framework of \cite{mac10}. The central idea of this
analysis is that the verb group is extracted from both sentences and
then infixed (at the place of the original verb group) in the first sentence.

\paragraph{Product rules} Some conjunctions have the simplest analysis
when we use the product formula. Look, for example, at the following sentence.

\exig.\label{s:prod}augmenter [np ses fonds propres ] [pp de 90 millions de
francs ] et [np les quasi-fonds propres ] [pp de 30 millions ]  \\
increase [np its equity {} ] [pp by 90 million {} francs ] and
[np its quasi-equity {} ] [pp by 30 million ]  \\

Here the verb ``augmenter'' (\emph{to augment}) takes both an $np$ and
a $pp$ argument.

We can derive these cases by assigning ``et'' the following formula. 
$$((np\bullet
pp)\backslash (np\bullet \Diamond_0\Box_0 pp))/(np\bullet pp)$$

The $\bullet I$ rule is easy to add to the chart parser. The
implementation is careful to use to product introduction rule only
when an adjacent chart item requires a product argument (a naive
implementation would concluded $A\bullet B$ from
\emph{any} adjacent chart items $A$ and $B$).

The elimination rules are more delicate and involve patterns such as
the following (these are easy to show valid using associativity of
$\Diamond_0\Box_0 C$).

$$
\begin{array}{ccc}
\infer[\textit{prod\_c}]{A\bullet\Diamond_0\Box_0 C}{A/B & B\bullet\Diamond_0\Box_0 C}
 &&
\infer[\textit{prod\_e}]{A}{(A/C)\bullet \Diamond_0\Box_0 C} \\
\end{array}
$$

Together, these allow us to combine $((np\backslash s)/pp)/np$ with
$np\bullet\Diamond_0\Box_0 pp$ as follows.

$$
\infer[\textit{prod\_e}]{np\backslash s}{\infer[\textit{prod\_c}]{((np\backslash s)/pp)\bullet
    \Diamond_0\Box_0 pp}{((np\backslash s)/pp)/np &
    np\bullet\Diamond_0\Box_0 pp}}
$$

\paragraph{Left-node raising} 
Very rarely, for a total of nine times in the entire corpus, we need
left-node raising, the symmetric operation of right-node raising. In the
example below, we have a conjunction of two combinations of two noun
post-modifiers $n\backslash n$: ``fran\c{c}ais A\'{e}rospatiale'' and ``italien Alenia''.

\exig.\label{lnr}... des groupes fran\c{c}ais A\'{e}rospatiale et
italien Alenia ... \\
... {of the} groups french A\'{e}rospatiale and italian Alenia ... \\
... of the french group A\'{e}rospatiale and italian (group) Alenia ...\\

By analysing ``et'' (and) as $((\Diamond_0\Box_0 n\backslash n)\backslash
(n\backslash n))/(\Diamond_0\Box_0 n\backslash n)$ we can use the
derivability of $n\backslash n, n\backslash n \vdash \Diamond_0\Box_0
n\backslash n$ as follows.

{\newcommand{\italien}{\textit{italien}}
\newcommand{\alenia}{\textit{Alenia}}
$$
\infer[\backslash I_1]{\italien \circ \alenia \vdash \Diamond_0\Box_0
n\backslash n}{\infer[\Diamond E_2]{x \circ (\italien \circ \alenia) \vdash
  n}{\infer[\textit{Hyp}]{x\vdash \Diamond_0\Box_0 n}{} & \infer[MA_l\Diamond_0]{\langle y \rangle^0 \circ (\italien
  \circ \alenia)\vdash n}{\infer[\backslash E]{(\langle y\rangle^0 \circ \italien)\circ
  \alenia\vdash n}{\infer[\backslash E]{\langle y\rangle^0 \circ
    \italien\vdash n}{\infer[\Box E]{\langle y\rangle^0 \vdash
      n}{\infer[\textit{Hyp}]{y\vdash \Box_0 n}{}} &\infer[\textit{Lex}]{n\backslash n}{\italien}} & \infer[L]{n\backslash n}{\alenia}}}}}
$$
}

\paragraph{Final implementation notes}

Since the final chart parser has many inference rules which apply only
in specific situations (essentially all rules, except for the basic AB
rules) and since the chart parser has a fair amount
of overhead trying (and failing) to match each of these rules, there
is a separate mechanism which verifies if the formulas contain any
patterns which trigger rules beyond the AB rules and if so, activate
all potentially useful rules. Therefore, the product rules are only
active if there is a formula of the form $A\bullet B$, the wrapping
rules only if there is a formula $A\backslash_1 A$, etc.

\section{Conclusion}

We have given a fairly high-level description of the multimodal
chart parser which is part of the type-logical treebank for
French. The source code, issued under the GNU Lesser General Public
License, contains much more detail.

\bibliographystyle{agsm}
\bibliography{moot}
\end{document}